\documentclass[sigconf]{acmart}
\usepackage{booktabs} 
\usepackage{soul}
\usepackage{graphicx}
\usepackage{subfigure}
\usepackage{colortbl}
\makeatletter
\newcommand{\tblcaption}[1]{\def\@captype{table}\caption{#1}}
\makeatother

\usepackage[skip=2pt]{caption}
\setcopyright{rightsretained}

\acmDOI{10.1145/nnnnnnn.nnnnnnn}

\acmISBN{978-x-xxxx-xxxx-x/YY/MM}

\acmConference[GECCO '19]{the Genetic and Evolutionary Computation Conference 2019}{July 13--17, 2019}{Prague, Czech Republic}
\acmYear{2019}
\copyrightyear{2019}

\acmArticle{4}
\acmPrice{15.00}

\begin{document}
\title{A Probabilistic Bitwise Genetic Algorithm for B-Spline based Image Deformation Estimation}

\author{Takumi Nakane}
\affiliation{%
  \institution{University of Fukui}
  \city{Fukui} 
  \country{Japan} 
}
\email{t-nakane@monju.fuis.u-fukui.ac.jp}

\author{Takuya Akashi}
\affiliation{%
  \institution{Iwate University}
  \city{Morioka} 
  \country{Japan} 
}
\email{akashi@iwate-u.ac.jp}

\author{Xuequan Lu}
\affiliation{%
  \institution{Deakin University}
  \city{Victoria} 
  \country{Australia} 
}
\email{xuequan.lu@deakin.edu.au}

\author{Chao Zhang}
\affiliation{%
  \institution{University of Fukui}
  \city{Fukui} 
  \country{Japan} 
}
\email{zhang@u-fukui.ac.jp}


\begin{abstract}
We propose a novel genetic algorithm to solve the image deformation estimation problem by preserving the genetic diversity. As a classical problem, there is always a trade-off between the complexity of deformation models and the difficulty of parameters search in image deformation. 
2D cubic B-spline surface is a highly free-form deformation model and is able to handle complex deformations such as fluid image distortions. However, 
it is challenging to estimate an apposite global solution.

To tackle this problem, we develop a genetic operation named \textit{probabilistic} \textit{bitwise} \textit{operation} (PBO) to replace the crossover and mutation operations, which can preserve the diversity during generation iteration and achieve better coverage ratio of the solution space. Furthermore, a selection strategy named annealing selection is proposed to control the convergence. Qualitative and quantitative results on synthetic data show the effectiveness of our method.
\end{abstract}

%
%



\maketitle

\section{Introduction}
Deformation estimation of an object between two images plays an essential role in various computer vision tasks. 
To model a complex non-rigid deformation, 2D Free-Form Deformation (FFD) using cubic B-spline is one of the classical models. 
To search the solution effectively, 
it is crucial to preserve genetic diversity during the population alternation \cite{hutter2002fitness}. However, the traditional GAs have been demonstrated to fall short for preserving diversity\cite{shir2005niching}, thus leading to undesired results. 

To alleviate the limitations, 
we propose a novel genetic operator called \textit{probabilistic bitwise operation} (abbreviated as PBO) to simultaneously play the roles of crossover and mutation. The main technical contributions 
are threefold. First, the genetic operation is introduced by inverting each bit with the probability determined with respect to the fitness value and bit order of each parameter. This ensures a population to better maintain the diversity and explore the solution space more comprehensively. Second, an annealing selection strategy for selecting PBO targets is proposed to control the convergence. Eventually, the movement of each control point is modeled to bridge the gap between the B-Spline deformation and the genetic framework, and a coarse-to-fine strategy using image pyramid is presented to 
robustly estimate enormous parameters. We show that the proposed method is capable of estimating complex image deformations.

\section{Probabilistic Bitwise Operation}
\begin{figure}[tb]
	\centering
	\includegraphics[width=0.75\linewidth]{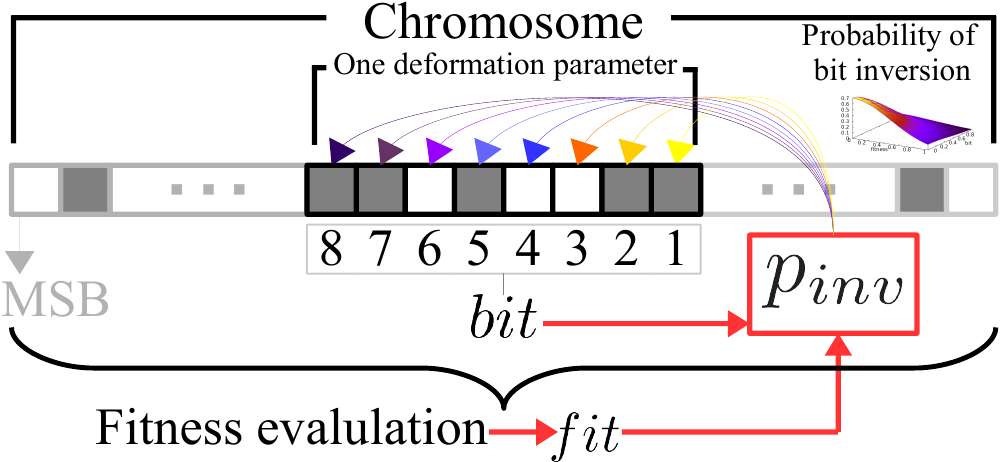}
    \caption{Illustration of PBO. Each deformation parameter is coded by a 8-bit (by default) binary code. Arrows in different color/intensity pointing to each bit represent the different inversion probability $P_{inv}$. Lower bit has higher $P_{inv}$.}
        \vspace{-1em}
    \label{fig:PBOIllustration}
\end{figure}

In a huge solution space, the search realized by the combination of crossover and mutation can hardly cover the whole space because of the decrease of diversity. The proposed PBO alleviates this problem through an independent genetic operation (i.e., individuals do not depend on others for updating). The probability of bit inversion $P_{inv}$ of some individual in a generation iteration relies only on the fitness value and bit order, as shown below.
\begin{equation}
 P_{inv} = w_{max}\exp{\left(-\frac{1}{2}\left(\frac{bit^2}{s^2_{bit}}+\frac{fit^2}{s^2_{fit}}\right)\right)},
 \label{equ:PBO}
\end{equation}
where $bit$ denotes the bit order over one parameter in the bit string, with the MSB defined as the left-most bit. $fit$ is the min-max normalized fitness value with respect to the current population. $s_{bit}$ and $s_{fit}$ are the parameters to control the smoothness of the distribution, and $w_{max}$ is a constant to determine the maximum probability that can be taken. PBO changes genes based on a simple motivation: the lower an individual is evaluated according to the fitness function, the greater its genes should be changed to explore further. The bitwise inversion can adjust the degree of change over a gene intuitively. We summarize the mechanism of PBO in Fig.\ref{fig:PBOIllustration}.

After performing PBO in the current generation, we extract the individuals of the next generation by roulette selection. It is worth pointing out that the selection probability named annealing selection rate $P_{ann}$ is proposed for selecting the individuals to conduct PBO. It is different from the roulette selection. Parameter $P_{ann}$, which is designed to decrease along with the progress of the generation iteration, plays a role on controlling convergence. In the $i$-th generation, it can be calculated as
\begin{equation}
 P_{ann} = \frac{1.0-\exp{\left(\frac{e}{G_{size}}i\right)}}{\exp{(e)}-1.0}(1.0-P_{min})+1.0,
 \label{equ:ann}
\end{equation}
where $e$ is a parameter adjusting the smoothness of the distribution, and $P_{min}$ is a constant indicating the value of $P_{ann}$ in the final generation $i = G_{size}$.

\section{Deformation Estimation}

\subsection{B-Spline Deformation Model}
We manipulate the source image with $W \times H$ pixels using control points and cubic B-spline. Control points are arranged in a lattice with intervals $\delta_x=W/(K-1)$ and $\delta_y=H/(L-1)$, where $K \times L$ is the number of control points which covers the source image. The deformation of a sample point $\vec{x}=(x,y)$ on the source image to $\vec{x}'$ is formulated by relating the displacement vector $\vec{d}_{i,j}$ with each surrounding control point $(i,j)$.
\begin{equation}
 \vec{x}' = \vec{x} + \sum_{l=0}^3 \sum_{m=0}^3 B_l(u)B_m(v)\vec{d}_{i+l,j+m},
\end{equation}
where $i=\lfloor x/\delta_x\rfloor -1$, $j=\lfloor y/\delta_y\rfloor -1$, $u=x/\delta_x-\lfloor x/\delta_x\rfloor$, $v=y/\delta_y-\lfloor y/\delta_y \rfloor$, and $B_l$, $B_u$ are the cubic B-spline basis functions. Since one sample point moves according to 16 neighboring control points, the deformation of the whole source image requires a total of $(k+2) \times (L+2)$ control points.

\subsection{Coarse-to-Fine Strategy}
\begin{figure}[tb]
	\centering
	\includegraphics[width=0.9\linewidth]{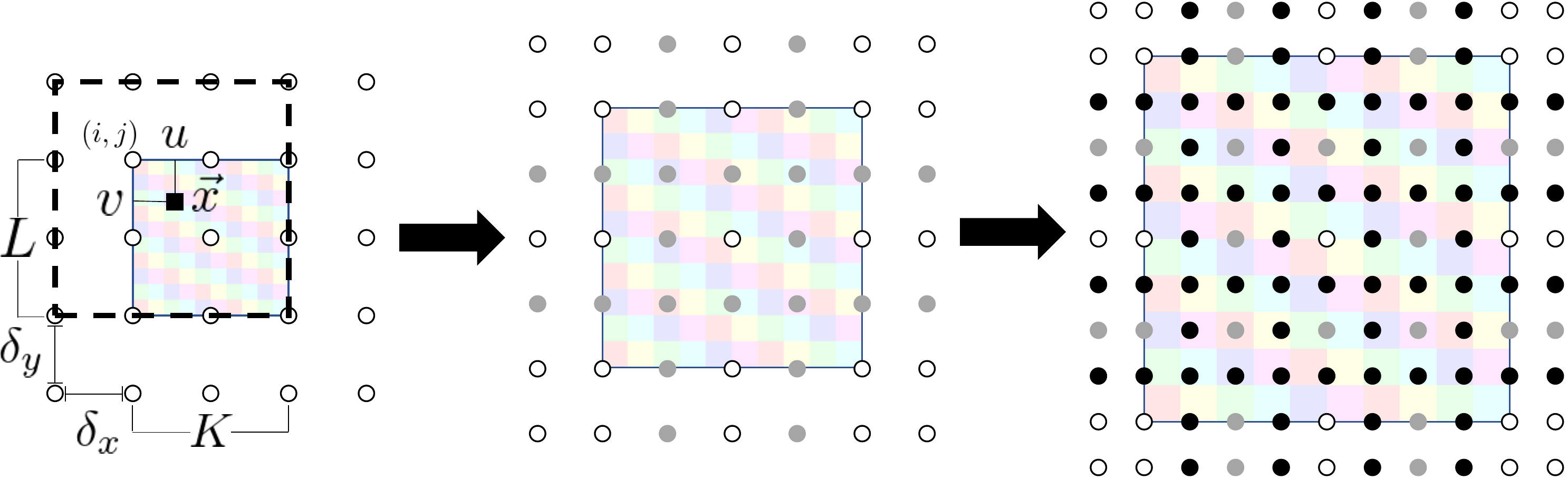}
    \caption{Top-3 levels of the B-spline deformation hierarchy. The parameters of control points of the previous level are inherited and the number of control points is increased.}
    \vspace{-1em}
    \label{fig:imagePyramid}
\end{figure}

The total number of parameters is  $(K+2) \times (L+2) \times 2$ with each displacement vector on the XY-plane holding two parameters. It is difficult to optimize all of them at the same time, as each parameter may affect others directly or indirectly. To settle this problem, we create image pyramids for both source and target images, shown in Fig.\ref{fig:imagePyramid}. The control points are then increased gradually along with the increase of image size in the pyramid. Specifically, the numbers of rows and columns are updated as $a_{n+1}=2a_n-1$, where $n$ is the level number from top to bottom. The generation iteration of the proposed method is performed in each hierarchy of the pyramid, and the initialization of current individuals is inherited from the result of the previous level.

\section{Experimental Results}
\begin{figure}[b]
 \def\@captype{table}
 \begin{minipage}[c]{.48\linewidth}
  \begin{center}
    \tblcaption{RMSE in pixels over 10 random deformations.}
   \begin{tabular}{p{0.4cm}>{\columncolor[gray]{0.7}}p{0.4cm}c>{\columncolor[gray]{0.7}}c} \hline
    \multicolumn{2}{c}{Image ID} & \multicolumn{2}{c}{RMSE} \\ \hline \hline
    1 & 6 & 3.35 & 4.48 \\
    2 & 7 & 2.92 & 4.63 \\
    3 & 8 & 3.61 & 4.24 \\
    4 & 9 & 4.19 & 3.62 \\
    5 & 10 & 3.62 & 5.10 \\ \hline
   \end{tabular}
   \label{table:result}
  \end{center}
 \end{minipage}
 \begin{minipage}[c]{.48\linewidth}
  \centering
      \vspace{-1.5em}
  \includegraphics[width=0.755\linewidth]{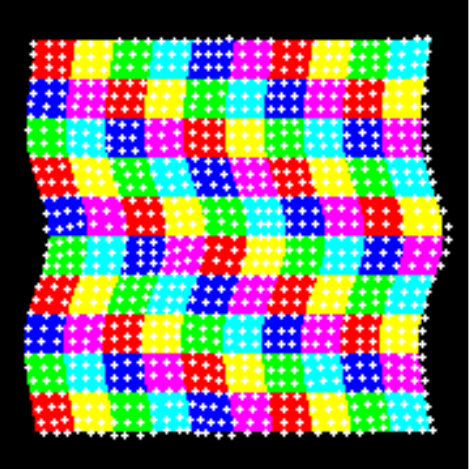}
  \caption{Points of the source are plotted on the target.}
  \label{fig:result}
 \end{minipage}
\end{figure}

We evaluate the effectiveness of our method on a synthetic dataset which consists of 10 types of random deformations. We create pyramids with 3-level hierarchy and conduct deformation using $K=L=3$ control points. The objective function is set to the sum of absolute difference of pixel intensities (i.e., gray scale value) between the source and target images. Each control point moves within a circle with a radius of 3 pixels, and its displacement vector is encoded with 5 bits per parameter. For quantitative analysis, we evaluate the deformation performed by the elite individual using the RMSE with respect to the manually selected points in the source image. The quantitative results are presented in Tab.\ref{table:result}. Fig.\ref{fig:result} demonstrates a qualitative example of result.

\section{Conclusions}
In this paper, we proposed a novel GA, which comprises of the \textit{probabilistic bitwise operation} (PBO) and the annealing selection, to solve the deformation estimation problem by preserving genetic diversity. This proposed GA estimates parameters of 2D cubic B-spline model to reconstruct the deformation between the source and target images, in a coarse-to-fine way. Both the qualitative and quantitative results on various synthetic data validate our method. As the future work, additional efforts would be made to deal with more complex deformations.

\bibliographystyle{ACM-Reference-Format}
\bibliography{GECCO2019-bibliography} 

\end{document}